# Intersectional Bias in Hate Speech and Abusive Language Datasets

**Jae Yeon Kim**\*, **Carlos Ortiz, Sarah Nam, Sarah Santiago, Vivek Datta**
University of California, Berkeley
Berkeley, California, USA

### Abstract

Algorithms are widely applied to detect hate speech and abusive language in social media. We investigated whether the human-annotated data used to train these algorithms are biased. We utilized a publicly available annotated Twitter dataset (Founta et al. 2018) and classified the racial, gender, and party identification dimensions of 99,996 tweets. The results showed that African American tweets were up to 3.7 times more likely to be labeled as abusive, and African American male tweets were up to 77% more likely to be labeled as hateful compared to the others. These patterns were statistically significant and robust even when party identification was added as a control variable. This study provides the first systematic evidence on intersectional bias in datasets of hate speech and abusive language.

## Introduction

Algorithms are widely applied to detect hate speech and abusive language in popular social media platforms such as YouTube, Facebook, Instagram, and Twitter. Using algorithms helps identify, at scale, which posts contain socially undesirable content. This computational method is efficient but not perfect. Most algorithms are trained with labeled data. What if the training data, used to detect bias in social media, were itself biased? Then, we would be in a situation where the algorithms that detect hate speech and abusive language, developed to prevent harm to protected groups such as racial minorities and women, exacerbate existing social disparities (Citron and Pasquale 2014; O'neil 2016).

We utilized a publicly available annotated Twitter dataset (Founta et al. 2018) and classified the racial, gender, and party identification dimensions of 99,996 tweets. The results showed that African American tweets were up to 3.7 times more likely to be labeled as abusive, and African American male tweets were up to 77% more likely to be labeled as hateful compared to the others. These patterns were statistically significant and robust even when party identification was added as a control variable. This study has many shortcomings. First and foremost, the evidence is suggestive because it is associational. Second, the magnitude of the intersectional bias is small and could be sensitive to measurement bias. Notwithstanding these caveats, this study is valuable because it provides the first systematic evidence on intersectional bias in datasets of hate speech and abusive language. Practitioners need to pay greater attention to the various forms of bias embedded in these datasets to avoid reinforcing a socially constructed stereotype, such as the presumed criminality of black men.

## Related Work

The unique contribution of this study lies in its intersectional angle. A robust body of work exists on the bias in datasets of hate speech and abusive language. Nevertheless, these studies examine this problem either from an exclusively racial (Waseem 2016; Waseem and Hovy 2016; Davidson, Bhattacharya, and Weber 2019; Sap et al. 2019) or gender-bias perspective (Tatman 2017; Park, Shin, and Fung 2018; Dixon et al. 2018). We build upon these works but also go one step further by looking at how the intersection of racial and gender bias matters. Highlighting the intersectional bias could be relevant because social science literature broadly emphasizes how the media portrays black men as threatening, hateful, and presumed criminals (Oliver 2003; Mastro, Blecha, and Seate 2015; Kappeler and Potter 2017) and how such media frames influence police interactions (Najdowski, Bottoms, and Goff 2015; Hall, Hall, and Perry 2016) and policy preference formation (Skinner and Hass 2016).

## Research Design

For transparency and reproducibility, we only used publicly available datasets. The annotated Twitter dataset (N = 99,996)[1] on hate speech and abusive language was created by a crowd-sourcing platform and its quality has been ensured by several rounds of validations. Founta et al., who generated the aforementioned dataset, defined abusive language as "any strongly impolite, rude, or hurtful language using profanity" and hate speech as "language used to express hatred towards a targeted individual or group" (5). We followed their definitions.

---

\*Corresponding author. PhD candidate in political science and graduate fellow at D-Lab and Data Science Education Program, University of California, Berkeley. jaeyeonkim@berkeley.edu

[1]In the data wrangling process, we discovered that 8,045 tweets were duplicates and removed them. Consequently, the size of the final dataset was reduced to 91,951 tweets.

The Twitter dataset does not contain any information on the racial, gender, or partisan identities of the authors of these tweets. We utilized additional public datasets to classify and fill in the missing information. Our primary interest was the interaction between race (defined as black or white) and gender (defined as male or female) and whether that generated a biased distribution of hateful and abusive labels. Such an underlying data distribution would generate uneven false positive and negative rates for different groups. However, human annotators could be biased not only in terms of race and gender but also in terms of political affiliation. This is likely true if annotators were recruited in the United States, where political polarization is extreme (Sides and Hopkins 2015; Broockman 2016). For this reason, we also classified party identification, the degree to which a person identifies with a particular political party.[2] To be clear, what we classified were not the actual racial, gender, or partisan identities of the authors of these tweets. The objective was to classify whether the language features expressed in a tweet were closer to the ones commonly expressed by one racial/gender/partisan group than those of other groups. To classify race, we leveraged African American and White English dialectal variations based on the model developed by Blodgett, Green, and O'Connor. This model matched 59.2 million geolocated Tweets from 2.8 million users with U.S. Census data. Gender and party identification classification were both based on the labeled data available at Kaggle's website. The gender data was originally provided by the Data for Everyone Library on CrowdFlower.[3] The party identification data was based on the tweets related to the 2018 US Congressional Election.[4]

For consistency, we applied identical preprocessing and feature extraction techniques to the tweets. We removed special characters from the tokenized tweets, turned them into lowercase, and reduced inflected words to their base forms using the Lancaster stemming algorithm (Paice 1990). We transformed these tokens into a document-term matrix using the bag-of-words model and constructing an n-gram with a maximum length of two. Then, we trained and tested the least absolute shrinkage and selection operator (Lasso) (Tibshirani 1996), naive Bayes (Maron 1961), and extreme gradient boosting (XGBoost) algorithms (Chen and Guestrin 2016). In the process, we divided 80% of the data into the training set and the rest of the data into the test set. Afterward, we measured the performance of each classifier using accuracy, precision, and recall scores. Table 1 summarizes the performances of these classifiers and shows Lasso outperformed the other classifiers.

[2]Party identification is different from political ideology, which demands more sophisticated political knowledge (Converse 1964).

[3]For more information, see https://www.kaggle.com/crowdflower/twitter-user-gender-classification

[4]For more information, see https://www.kaggle.com/kapastor/democratvsrepublicantweets

Table 1: Classifier performance

| Models | Accuracy | Precision | Recall | Label |
|---|---|---|---|---|
| Lasso | 0.69 | 0.68 | 0.73 | Male |
| Bayes | 0.62 | 0.58 | 0.86 | Male |
| XGBoost | 0.65 | 0.65 | 0.65 | Male |
| Lasso | 0.73 | 0.73 | 0.77 | Female |
| Bayes | 0.69 | 0.65 | 0.84 | Female |
| XGBoost | 0.71 | 0.72 | 0.71 | Female |
| Lasso | 0.73 | 0.72 | 0.71 | Party ID |
| Bayes | 0.71 | 0.72 | 0.66 | Party ID |
| XGBoost | 0.70 | 0.70 | 0.67 | Party ID |

The data analysis was correlational and thus, descriptive. We first described the bivariate relationship between race and gender and then added uncertainty about the measures using bootstrapping (Efron and Tibshirani 1986). We further investigated how the interaction between race and gender influences the distribution of hateful and abusive labels using logistic regression models. By taking a statistical modeling approach, we estimated the partial effect of the interaction bias on the outcomes while controlling for partisan bias.

## Hypotheses

- **Racial bias**: The first hypothesis is about between-group differences. Consistent with the prior research, we expect that tweets more closely associated with African American than White English language features would be more likely to be labeled as abusive and hateful.

- **Intersectional bias:** The second hypothesis is about within-group differences. Influenced by broad social science literature, we argue that tweets more closely associated with African American males than other groups' language features are more likely to be labeled as hateful.

## Results

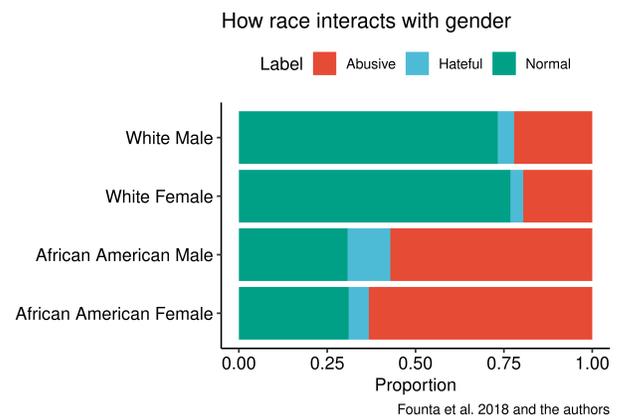

Figure 1: Descriptive analysis

To begin with, Figure 1 displays the bivariate relationship between tweets classified by race and by gender. The X-

axis indicates the proportion of tweets labeled as either abusive, hateful, or normal within four intersectional group categories (e.g., African American male). The figure shows two patterns. First, African American tweets are more likely to be labeled as abusive than White tweets are. Second, African American male tweets are more likely to be labeled as hateful as compared to the other groups.

One limitation of Figure 1 is that it does not show the uncertainty of the measures. Figure 2 addresses this problem by randomly resampling the data 1,000 times with replacement and stratifying on race, gender, and label type. There are several differences between Figure 1 and Figure 2. The bootstrapping method produces 95% confidence intervals, indicated by the error bars in the figure. Another difference is that the Y-axis shows label types rather than intersectional group categories. This figure reaffirms what we found earlier: African American tweets are overwhelmingly more likely to be labeled as abusive than their White counterparts. An opposite pattern is found in the normal label; White tweets are far more likely to be labeled as normal than their African American counterparts. These patterns are statistically significant because they are far outside confidence intervals. Gender difference matters little in these cases. By contrast, the intersection between race and gender matters in hate speech annotation. African American male tweets are far more likely to be labeled as hateful than the rest of the groups are. African American female tweets are only slightly more likely to be labeled as hateful than their White counterparts are.

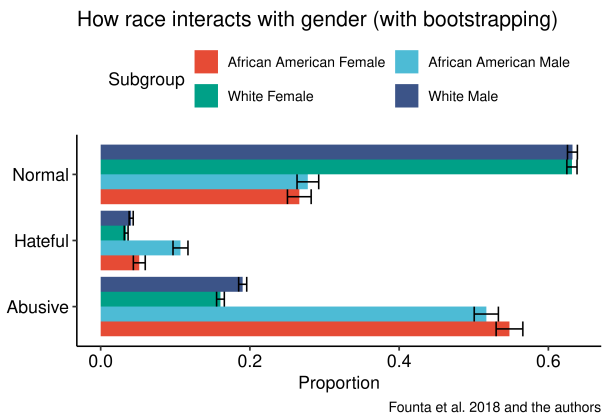

Figure 2: Bootstrapping results

Figure 3 extends the previous investigation by adding party identification as a control variable. We constructed two logistic regression models. In both models, the dependent variable was an abusive or hateful category defined as dummy variables (yes = 1, no = 0). The first model did not involve party controls and its predictor variables were race, gender, and their interaction. The second model involved party controls and its predictor variables were race, gender, party identification, the intersection between race and gender, and the intersection between and race and party identification. In the figure, the results of the first model are in-dicated by light blue, and the second model by red dots. The error bars indicate two standard errors (approximately 95% confidence intervals). The Y-axis indicates key predictor variables and the X-axis shows the likelihoods, which we calculated from the logistic regression estimates to help interpretation. The figure demonstrates that African American language features are most likely to be labeled as abusive, and African American male language features are most likely to be labeled as hateful. This pattern is robust across the two models. To be precise, if tweets were associated with African American language features, the likelihood of these tweets to be labeled as abusive increased by up to 3.7 times. If tweets were associated with African American male language features, the likelihood of these tweets to be labeled as hateful increased by up to 77%.

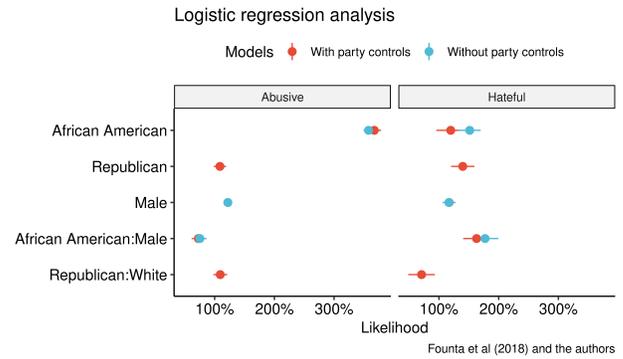

Figure 3: Logistic regression analysis

## Discussions and Conclusion

This study provides the first systematic evidence on intersectional bias in datasets of hate speech and abusive language. More importantly, the finding that African American men are closely associated with hate speech is consistent with broad social science research on the criminalization of African American men. The evidence emphasizes the importance of taking an interdisciplinary approach. The proliferation of machine learning applications is new. However, human biases have a much longer history and are broadly studied outside computer science. Social science knowledge could provide insights into the explicit and implicit bias embedded in datasets used in machine learning applications, including systems to detect hate speech and abusive language.

However, this study has many shortcomings. The first caveat is that the statistical model could be incomplete. The multivariate regression model is naive and likely to be underspecified. Missing variables in the model may cause selection bias. A better approach would be to design an experiment in which researchers could manipulate the source of biases—different types of language features—and directly examine their causal effects. Sap et al. showed how a behavioral experiment can be conducted for identifying racial bias in datasets of hate speech and abusive language. Experimental evidence for intersectional bias is still lacking and remains critical for future research.

The second caveat is that the data could be inaccurate. This problem is particularly concerning because the magnitude of the intersectional bias is small (see Figure 3). All of the key predictor variables were not directly observed but were based on machine-enabled text classification. Table 1 displays that these predictions show modest performance (between 69% and 73% accuracy scores). Uncertainty in the data may not destabilize inference if the effect size is large enough; an increase of up to 3.7 times due to racial bias is difficult to remove. By contrast, an increase of up to 77% due to intersectional bias could be sensitive to measurement error. The findings reported here should not be taken at their face value and should be followed up with further investigation.

# Author Contributions

Jae Yeon Kim is a project lead. Kim designed the research, collected and analyzed data, and wrote the paper. The rest of the authors are undergraduate research assistants. They helped with data collection and analysis. Their names are listed alphabetically.

# Additional Resources

All replication files are found at https://github.com/jaeyk/intersectional-bias-in-ml

# Acknowledgments

We are grateful to two anonymous reviewers for their constructive comments.